\documentclass[9pt,shortpaper,twoside,web]{ieeecolor2}
\usepackage{generic}
\usepackage{cite}
\usepackage{amsmath,amssymb,amsfonts}
\usepackage{algorithmic}
\usepackage{graphicx}
\usepackage{textcomp}
\usepackage{amsmath}
\usepackage{booktabs}
\usepackage{subfig}
\usepackage{multirow}
\usepackage{tabularx}
\usepackage{array}
\usepackage{xcolor}

\def\BibTeX{{\rm B\kern-.05em{\sc i\kern-.025em b}\kern-.08em
    T\kern-.1667em\lower.7ex\hbox{E}\kern-.125emX}}
\begin{document}
\title{A Novel Passive Occupational Shoulder Exoskeleton With Adjustable Peak Assistive Torque Angle For Overhead Tasks}
\author{Jin Tian, \IEEEmembership{Graduate Student Member, IEEE}, Haiqi Zhu, Changjia Lu, Chifu Yang, Yingjie Liu, Baichun Wei, and Chunzhi Yi, \IEEEmembership{Member, IEEE}
\thanks{This work was supported in part by the National Natural Science Foundation of China under Grant 62306083 and the Postdoctoral Science Foundation of Heilongjiang Province of China under Grant LBH-Z22175. (\textit{Correspondence author: Baichun Wei, Chunzhi Yi})}
\thanks{Jin Tian and Chifu Yang are with the School of Mechatronics Engineering, Harbin Institute of Technology, Harbin, 150001, China (e-mail: jin.tian@stu.hit.edu.cn; cfyang@hit.edu.cn).}
\thanks{Haiqi Zhu, Baichun Wei, and Chunzhi Yi are with the School of Medicine and Health, Harbin Institute of Technology, Harbin, 150001, China (e-mail: haiqizhu@hit.edu.cn; bcwei@hit.edu.cn; chunzhiyi@hit.edu.cn).}
\thanks{Changjia Lu and Yingjie Liu are with the Emergency Science Research Institute, Chinese Institute of Coal Science, Beijing, 100013, China (e-mail: lusipshan@163.com; liuyingjie@mail.ccri.ccteg.cn).}}

\maketitle

\begin{abstract}
\textit{Objective:} Overhead tasks are a primary inducement to work-related musculoskeletal disorders. Aiming to reduce shoulder physical loads, passive shoulder exoskeletons are increasingly prevalent in the industry due to their lightweight, affordability, and effectiveness. However, they can only accommodate a specific task and cannot effectively balance between compactness and sufficient range of motion. \textit{Method:} We proposed a novel passive occupational shoulder exoskeleton to handle various overhead tasks with different arm elevation angles and ensured a sufficient ROM while compactness. By formulating kinematic models and simulations, an ergonomic shoulder structure was developed. Then, we presented a torque generator equipped with an adjustable peak assistive torque angle to switch between low and high assistance phases through a passive clutch mechanism. Ten healthy participants were recruited to validate its functionality by performing the screwing task. \textit{Results:} Measured range of motion results demonstrated that the exoskeleton can ensure a sufficient ROM in both sagittal (164°) and horizontal (158°) flexion/extension movements. The experimental results of the screwing task showed that the exoskeleton could reduce muscle activation (up to 49.6$\%$), perceived effort and frustration, and provide an improved user experience (scored 79.7 out of 100). \textit{Conclusion:} These results indicate that the proposed exoskeleton can guarantee natural movements and provide efficient assistance during overhead work, and thus have the potential to reduce the risk of musculoskeletal disorders. \textit{Significance:} The proposed exoskeleton provides insights into multi-task adaptability and efficient assistance, highlighting the potential for expanding the application of exoskeletons.
\end{abstract}


\renewcommand{\thefootnote}{} 
\footnotetext[0]{Copyright (c) 2021 IEEE. Personal use of this material is permitted. However, permission to use this material for any other purposes must be obtained from the IEEE by sending an email to pubs-permissions@ieee.org.} 

\section{Introduction}
\label{sec:introduction}
\IEEEPARstart{W}{ORK-RELATED} musculoskeletal disorders (WMSDs) are the main health threats to workers in China, Europe, and even the world \cite{cieza2020global}, \cite{enoka2008muscle}. Shoulder WMSDs have received widespread attention due to their high prevalence (13$\%$ of cases) and prolonged absenteeism (23 days) \cite{wang2017work}. Overhead tasks, often related to extreme and awkward shoulder postures, are defined as working with hands above the shoulder height \cite{sluiter2001criteria}.They have been recognized as a primary contributor to shoulder WMSDs \cite{ding2023lightweight}. Although robots have been involved in many factory tasks, numerous complex activities that demand human cognition or operate in narrow spaces still require workers, especially in the construction industry \cite{huysamen2018evaluation}. Occupational shoulder exoskeleton (OSE) technology is an available solution to reduce the burden on the shoulder joint and the incidence of its WMSDs \cite{park2021shoulder}.

OSEs can be classified into active and passive categories. In general, active OSEs can provide higher, more precise, and versatile assistive torques \cite{zhou2021kinematics, grazi2022kinematics, otten2018evaluation}. However, they have larger volumes, resulting in bulkier and heavier, and are prone to malfunctions (i.e., increased maintenance costs) and collision \cite{kim2022passive}. Passive OSEs (POSEs) characterize as cheaper, lightweight design and a trade-off between effectiveness and wearability \cite{grazi2020design} \cite{pacifico2022exoskeletons}, thus are increasingly accepted by several manufacturers in real-application scenarios. They can provide arm gravity compensation during overhead works without external power sources. Recent studies have demonstrated that wearing POSEs can reduce muscle activation during overhead tasks performed in laboratory and field settings \cite{de2020passive, de2023passive, iranzo2020ergonomics}. Overhead tasks relate to different positions, especially different elevation angles of shoulder (e.g., screwing task of the automobile manufacturing industry \cite{pinho2022shoulder, yin2020effects, kim2018assessing}, aircraft squeeze riveting and sealing tasks \cite{jorgensen2022influence, jorgensen2022impact}, the welding task of shipbuilding \cite{hyun2019light}, order picking task \cite{de2020passive} and so on). Even though POSEs have been commercialized, some pending issues still remain, thus impede the user acceptance and assistive efficiency \cite{rossini2021design}.

One of the main issues is the insufficient kinematic compatibility between the exoskeleton and the user [23]. POSEs can be classified into two types based on whether there is an exoskeletal shoulder structure above the human shoulder \cite{kim2022passive}. Exoskeletons with a rigid structure above the human shoulder, such as the ShoulderX \cite{pinho2022shoulder} and Airframe \cite{mcfarland2022level}, typically feature a degree of freedom (DoF) next to the shoulder to follow shoulder flexion/extension (F/E) movements. These exoskeletons are compact and thus are suitable for performing overhead tasks in narrow spaces. However, they may induce collisions between the shoulder and the exoskeleton at high elevation angles, resulting in an insufficient range of motion (ROM). Exoskeletons without a structure above the shoulder, such as the MATE \cite{pacifico2020experimental}, Exo4Work \cite{de2022occupational}, and H-PULSE \cite{grazi2022kinematics, grazi2020design}, avoid collisions at high elevation angles. However, they generally incorporate multi-DoF structures in the back to accommodate shoulder movements, which leads to a larger volume. The drawback may result in collisions with the surrounding objects in narrow spaces. In existing exoskeletons, achieving both sufficient ROM and compactness is mutually exclusive. To achieve a balance between them, Junsoo Kim et al. \cite{kim2022passive} designed an exoskeleton with tilted shoulder structures for assisting overhead tasks. However, it still cannot provide a sufficient ROM for horizontal F/E movements, which may induce a decreased user experience. There is an urgent need for an exoskeleton that can simultaneously provide sufficient ROM and compactness.

Another main issue of current POSEs is the task-specific or individual-specific design, corresponding to a fixed assistive torque profile and thus a fixed assistive region (i.e. the subset of ROM where the exoskeleton can provide assistance), which may not be optimal for other tasks \cite{maurice2019objective, hyun2019light}. Considering that the assistive torque profile for overhead tasks are mainly inverse U-shaped due to the need of compensating for the gravity of arm elevation and lowering, the torque profile should be adaptive to tasks and subjects, essentially characterized by the amplitude of the peak torque and the timing of peak torque (i.e. the peak assistive torque angle, PATA). Different overhead tasks correspond to different postures and PATAs, thus require different optimal torque profiles \cite{grazi2020design, pacifico2020experimental}. In addition, the same task may have different PATAs for individuals of different heights, which requires providing personalized torque profiles \cite{vazzoler2022analysis}. The amplitude of the peak assistive torque can be easily tuned by adjusting the extension length of the spring. For example, the MATE can manually change the peak torque by a knob \cite{pacifico2020experimental}, and the H-PULSE can adjust the spring length by a spindle drive coupled with a servomotor \cite{grazi2022kinematics, grazi2020design}. The timing of the peak assistive torque ensures that the POSE can provide timely assistance when the shoulder torque reaches its peak value at some specific elevation angles of shoulder \cite{rossini2021design}. Fig. 1 depicts the torque profiles when the PATA matches or mismatches the requirement of the task in an overhead task with a target angle of 120º. When the PATA matches the target angle of the task, it ensures timely assistance and reduces energy loss, making it a more efficient assistive strategy. However, current POSEs can only change the PATA through reconstruction, which means they cannot provide torque profiles for different tasks or individuals, i.e., insufficient adaptability \cite{hyun2019light, vazzoler2022analysis}. In summary, different tasks and individuals should be matched with different PATA, lacking in current POSEs. Therefore, a POSE with adjustable PATA is needed to enhance its task and subject adaptability.

Inspired by the issues mentioned above, the object of this study is to develop, implement, and preliminarily evaluate a novel POSE for overhead tasks, named HIT-POSE. Our prototype can provide sufficient ROM of the human shoulder and possess adjustable PATA, thus adapting to various overhead tasks and individuals. The major contributions of our work are as follows:

(a) We designed a torque generator with adjustable PATA to adapt to different overhead tasks and individuals. To the best of our knowledge, it’s the first torque generator that can provide adjustable PATA to achieve more efficient assistance.

(b) We developed an ergonomic shoulder structure through kinematic modeling and simulation, which can guarantee sufficient ROM for the shoulder while remaining compact (without redundant structures in the back), thus reducing the risk of collisions with the environment.

(c) We also investigated how the modulation of the PATA could affect objective and subjective evaluation indicators in overhead tasks via experimental assessment, thus validating the effectiveness of the exoskeleton.

The article is organized as follows. Section \ref{sec2} introduces the structure design of HIT-POSE in detail. Section \ref{sec3} presents the experimental evaluation methods. The evaluation results are reported in Section \ref{sec4}. Section \ref{sec5} demonstrates the discussion of the results, followed by a conclusion of the full article.

\begin{figure}[!t]
	\centerline{\includegraphics[width=\columnwidth]{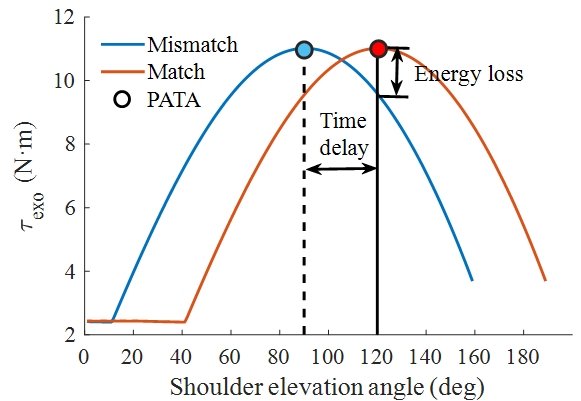}}
	\caption{The torque profiles when the PATA matches (orange line) or mismatches (blue line) the target angle of the task in an overhead task with a target angle of 120º.}
	\label{fig1}
\end{figure}

\section{System Design of HIT-POSE}
\label{sec2}
\subsection{System Overview}

The HIT-POSE is designed to compensate for shoulder gravitational torque while providing sufficient ROM for the shoulder, as illustrated in Fig. 2. The total weight of the device is 2.8 kg. The HIT-POSE comprises four main modules: (a) the ergonomic shoulder structure, (b) the torque generator with adjustable PATA, (c) the physical user-exoskeleton interface (PUEI), and (d) the size regulation module. The guiding principles of design are safety, compactness, efficiency, adequate ROM, and wearability. HIT-POSE can adjust the PATA and the peak torque to generate more versatile and efficient torque profiles, which can effectively assist the shoulder in multiple overhead tasks with different levels of arm elevations.

\begin{figure}[!t]
	\centerline{\includegraphics[width=\columnwidth]{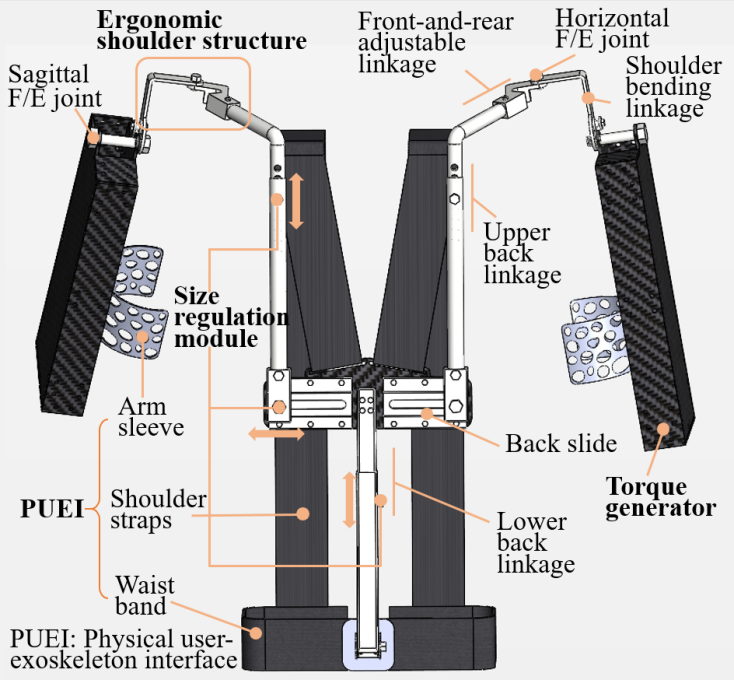}}
	\caption{An overview of HIT-POSE.}
	\label{fig2}
\end{figure}

\subsection{Ergonomic Shoulder Structure}
In this subsection, we conducted kinematic modeling of the exoskeleton and the user and simulated ROM to select optimal design parameters, and therefore ensure sufficient ROM for the shoulder when wearing the exoskeleton. The shoulder bending linkage is connected to the torque generator and front-and-rear adjustable linkage through the sagittal and horizontal F/E joints, respectively (as shown in Fig. 2). The proposed exoskeleton is a POSE with a horizontal F/E joint above the shoulder and without redundant structures in the back. Our object here is to provide sufficient ROM for the shoulder, especially at high elevation angles.

\begin{figure}[!t]
	\centerline{\includegraphics[width=\columnwidth]{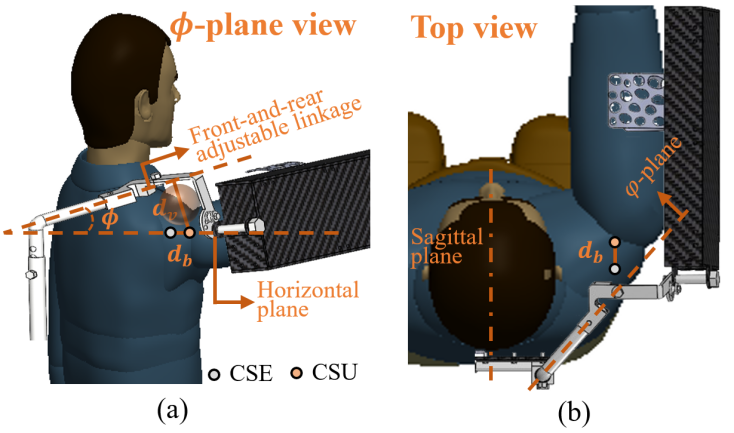}}
	\caption{Definition of the design parameters (\(\phi\), \(d_{v}\), \(d_{b}\)). (a) \(\phi\) and \(d_{v}\) are defined from the \(\phi\)-plane view, the orange clear orb represents the room between the CSU and the front-and-rear adjustable linkage. (b) \(d_{b}\) is defined from the top view.}
	\label{fig3}
\end{figure}

The intersection of the rotational axes of the sagittal and horizontal F/E joints is considered the center of the shoulder of the exoskeleton (CSE), as shown in Fig. \ref{fig3}. The CSE is close to the center of the shoulder of the user (CSU), enabling the torque generator to trace the shoulder movements with a relatively low DoF (Fig. \ref{fig3}). However, it does not coincide with the CSU, indicating the necessity to find the optimal design parameters. As explained in \cite{ludewig2009motion} and \cite{stokdijk2000glenohumeral}, the shoulder motion can be observed with humeral elevation and posterior displacement of the CSU. This phenomenon introduces great design inspiration to us.

Fig. \ref{fig3} demonstrates the extracted design parameters (\(\phi\), \(d_{v}\), \(d_{b}\)), which can reconfigure the shoulder structure. We defined the \(\phi\)-plane as the plane containing the front-and-rear adjustable linkage and perpendicular to the horizontal plane (Fig. \ref{fig3}). Viewed from the \(\phi\)-plane, the pitch angle (\(\phi\)) is the angle between the front-and-rear adjustable linkage and horizontal plane, and the vertical distance (\(d_{v}\)) refers to the perpendicular distance from the CSU to the front-and-rear adjustable linkage, as presented in Fig. \ref{fig3}a. Bias distance \(d_{b}\) is defined as the distance between the CSE and CSU from the top view (Fig. \ref{fig3}b). \(\phi\) and \(d_{v}\) can determine the configuration of the front-and-rear adjustable linkage and shoulder bending linkage, respectively. \(d_{b}\) can be regulated by adjusting the length of the front-and-rear adjustable linkage, to adapt to different users and ensure comfort. The orange clear orb (Fig. \ref{fig3}a) can be enlarged by increasing \(\phi\) and \(d_{v}\), provided for humeral elevation and avoiding collision at high elevation angles. Increasing \(d_{b}\) shifts CSE posteriorly, compensating for the displacement of the CSU during shoulder movement. Overall, a proper combination of design parameters (\(\phi\), \(d_{v}\), \(d_{b}\)) can increase the user’s ROM when wearing the exoskeleton.

\begin{figure}[!t]
	\centerline{\includegraphics[width=\columnwidth]{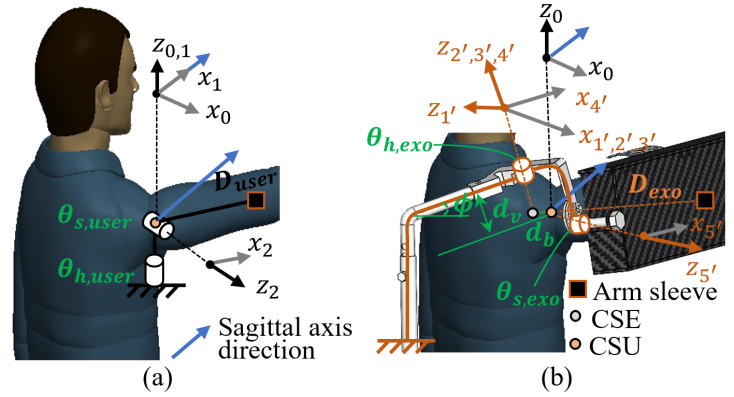}}
	\caption{ Kinematic models. (a) The user. (b) The exoskeleton.}
	\label{fig4}
\end{figure}

\begin{table}
	\centering
	\caption{DH PARAMETERS OF THE USER}
	\setlength{\tabcolsep}{3pt}
	\begin{tabular}{p{40pt} p{35pt} p{35pt} p{35pt}  p{45pt} ll}
		\toprule
		& 
		\(\alpha_{i-1}\)& \(a_{i-1}\)&\(d_{i}\)&\(\theta_{i}\) \\
		\midrule
		1’(\(\theta_{h,user}\))& 0& 0& 0& $\pi$/2+\(\theta_{h,user}\) \\
		2’(\(\theta_{s,user}\))& -$\pi$/2& 0& 0& -$\pi$/2+\(\theta_{s,user}\) \\	
		Arm sleeve& 0& \(D_{user}\)& 0&	0 \\
		\bottomrule
	\end{tabular}
	\label{tab1}
\end{table}

\begin{table}
	\centering
	\caption{DH PARAMETERS OF THE EXOSKELETON}
	\setlength{\tabcolsep}{3pt}
	\begin{tabular}{p{40pt} p{35pt} p{35pt} p{35pt}  p{50pt} ll}
		\toprule
		& 
		\(\alpha_{i-1}\)& \(a_{i-1}\)&\(d_{i}\)&\(\theta_{i}\) \\
		\midrule
		1(\(d_{b}\))& $\pi$/2& 0& \(d_{b}\)& 0 \\
		2(\(\phi\), \(d_{v}\))& -$\pi$/2+\(\phi\)& 0& \(d_{v}\)& 0 \\	
		3(\(\theta_{h,exo}\))& 0& 0& 0&	$\pi$/2+\(\theta_{h,user}\) \\
		4(\(d_{v}\))& 0& 0& -\(d_{v}\)& 0 \\
		5(\(\theta_{s,exo}\))& -$\pi$/2& 0& 0& -\(\phi\)-$\pi$/2+\(\theta_{s,exo}\) \\
		Arm sleeve& 0& \(D_{exo}\)& 0& 0& \\
		\bottomrule
	\end{tabular}
	\label{tab2}
\end{table}

\begin{table}
	\centering
	\setlength{\tabcolsep}{3pt}
	\begin{tabular}{p{50pt} p{185pt} ll}
		\toprule
		\textbf{Simulation $\mathrm{\uppercase\expandafter{\romannumeral1}}$}& Calculate the user’s ROM when wearing the exoskeleton \\
		\midrule
		\textbf{Input}& \(\theta_{h,exo}\), \(\theta_{s,exo}\), \(\phi\), \(d_{v}\), \(d_{b}\), \(D_{exo}\), \(D_{user}\) \\
		\textbf{Output}& ROM simulation results \\
		1:& Calculating the position of the arm sleeve (\(p_{x}\), \(p_{y}\), \(p_{z}\)) via forward kinematics of the exoskeleton; \\
		2:& Solving \(\theta_{h,user}\) and \(\theta_{s,user}\) using inverse kinematics of the user and position of arm sleeve (\(p_{x}\), \(p_{y}\), \(p_{z}\));\\
		3:& Checking collision between the exoskeleton and user;\\
		4:& Obtain the non-collisional joint angles of the user.\\
		\bottomrule
	\end{tabular}
	\label{sim1}
\end{table}

To better explore the effect of design parameters on ROM, the kinematic analyses were conducted, as shown in Fig. \ref{fig4}. In this study, we only consider 2-DoFs of the user’s shoulder, a horizontal and sagittal F/E DoF, whose angles were denoted by \(\theta_{h,user}\) and \(\theta_{s,user}\) (Fig. \ref{fig4}a). Similarly, the exoskeleton’s shoulder has 2-DoFs, whose angles are denoted by \(\theta_{h,exo}\) and \(\theta_{s,exo}\) (Fig. \ref{fig4}b). These four angles represent the ROM for both the human and the exoskeleton. \(D_{user}\) and \(D_{exo}\) represent the distances from the CSU and CSE to the arm sleeve (Fig. \ref{fig4}), respectively.

Tables \ref{tab1} and \ref{tab2} demonstrate the specific Denavit-Hartenberg (DH) parameters of both models. The association between the two models is that the coordinate origin (the CSU in Fig. \ref{fig4}) and arm sleeve are in the same location. Based on the characteristics, we formulated a simulation pipeline via the analytical method to calculate the user’s ROM when wearing the exoskeleton, as displayed in Simulation $\mathrm{\uppercase\expandafter{\romannumeral1}}$. Notably, the principle for checking collisions is that the angles of the user cannot be greater than that of the exoskeleton.

\begin{figure}[!t]
	\centerline{\includegraphics[width=\columnwidth]{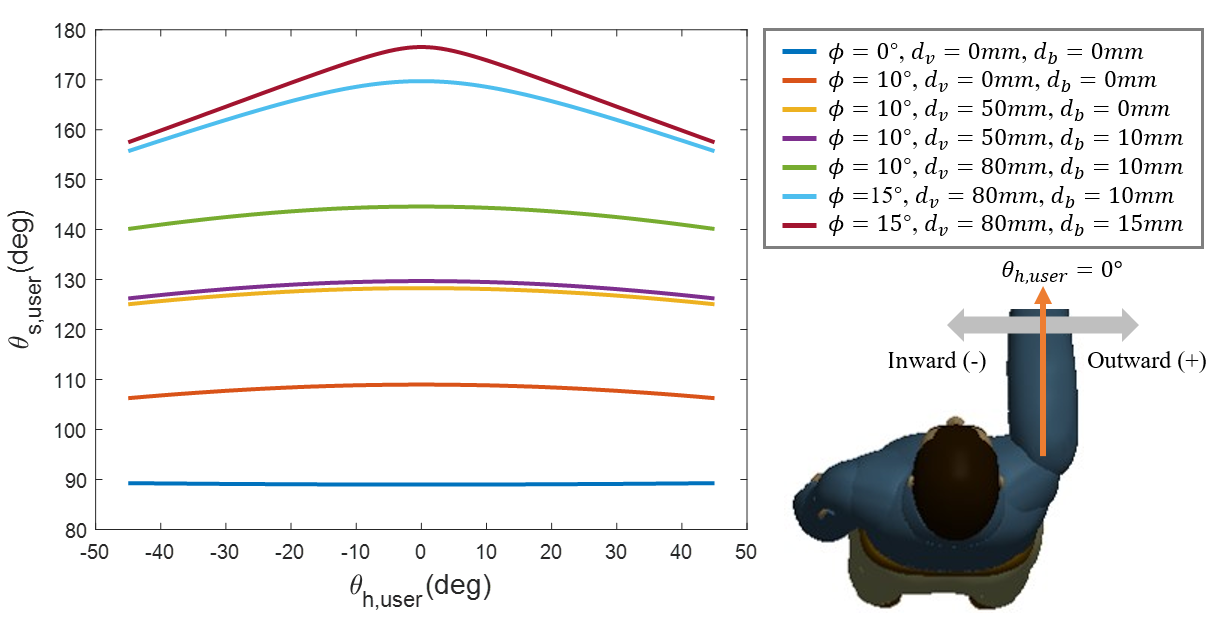}}
	\caption{ROM simulation results of the user with seven sets of design parameters (\(\phi\), \(d_{v}\), \(d_{b}\)).}
	\label{fig5}
\end{figure}

A previous study has stated that the CSU is about 2 cm below the acromion \cite{liu2007estimation}. Additionally, the CSU is vertically displaced by around 6 cm during shoulder movement \cite{nef2009armin}. Such prior knowledge provides a basis for choosing the value of \(d_{v}\). We provide seven sets of design parameters and their ROM simulation results, as shown in Fig. \ref{fig5}. The area below each set of curves is the corresponding ROM. When \(\theta_{h,user}\)=0, the maximum value of \(\theta_{s,user}\) is obtained. The ROM can be enlarged by increasing \(\phi\) and \(d_{v}\), thus preventing collisions at high elevation angles. To maximize the user’s ROM, \(\phi\) and \(d_{v}\) were set to 15º and 80 mm, respectively. As \(d_{b}\) increases, the CSE moves toward the posterior of the user, causing increased ROM, but too large \(d_{b}\) will lead to discomfort. Eventually, we take the penultimate set (\(\phi\)=15°, \(d_{v}\)=80mm, \(d_{b}\)=10mm) as the final design parameters. In this set, the ROM is sufficient to ensure the user’s natural movement and complete overhead tasks.

\begin{figure*}[!t]
	\centering
	\subfloat[]{\includegraphics[width=0.39\textwidth]{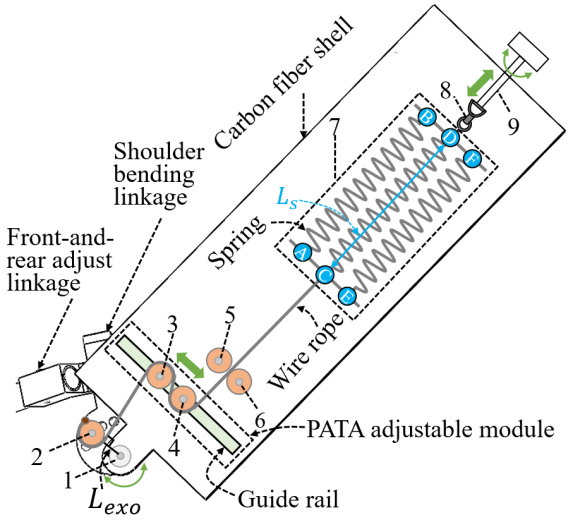}\label{fig6a}}
	\hfill
	\subfloat[]{\includegraphics[width=0.30\textwidth]{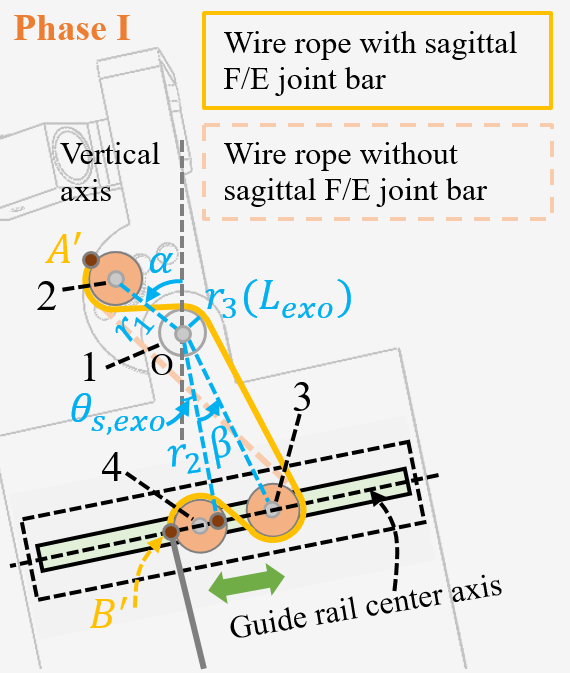}\label{fig6b}}
	\hfill
	\subfloat[]{\includegraphics[width=0.295\textwidth]{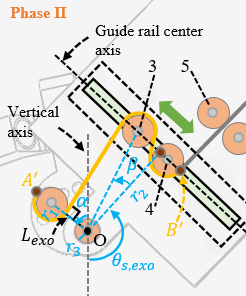}\label{fig6c}}
	\caption{The overall structure and design parameters of the torque generator with adjustable PATA. (a) Components of the torque generator. (b) Torque generator with parameters (\(\alpha\), \(\beta\), \(r_{1}\), \(r_{2}\), \(r_{3}\)) after the wire rope is attached to the sagittal F/E joint bar (Phase $\mathrm{\uppercase\expandafter{\romannumeral1}}$). (c) Torque generator with parameters (\(\alpha\), \(\beta\), \(r_{1}\), \(r_{2}\), \(r_{3}\)) before the wire rope is attached to the sagittal F/E joint bar (Phase $\mathrm{\uppercase\expandafter{\romannumeral2}}$). 1: Sagittal F/E joint bar. 2: Pulley fixed to the shoulder bending linkage. 3-6: Pulleys fixed to the carbon fiber shell. 7: Parallel spring group. 8: 8-shaped loop. 9: Spring pretension tuning module. \(L_{exo}\): the force arm of the assistive torque.}
	\label{fig6}

\end{figure*}

\begin{figure*}[!t]
	\subfloat[]{\includegraphics[width=0.30\textwidth]{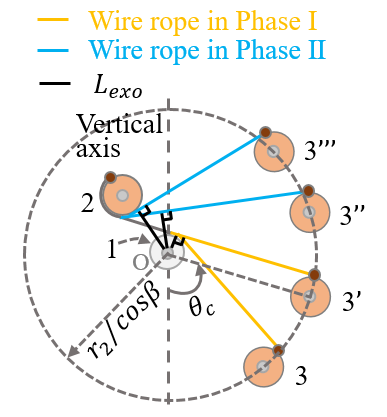}\label{fig7a}}
	\hfill
	\subfloat[]{\includegraphics[width=0.60\textwidth]{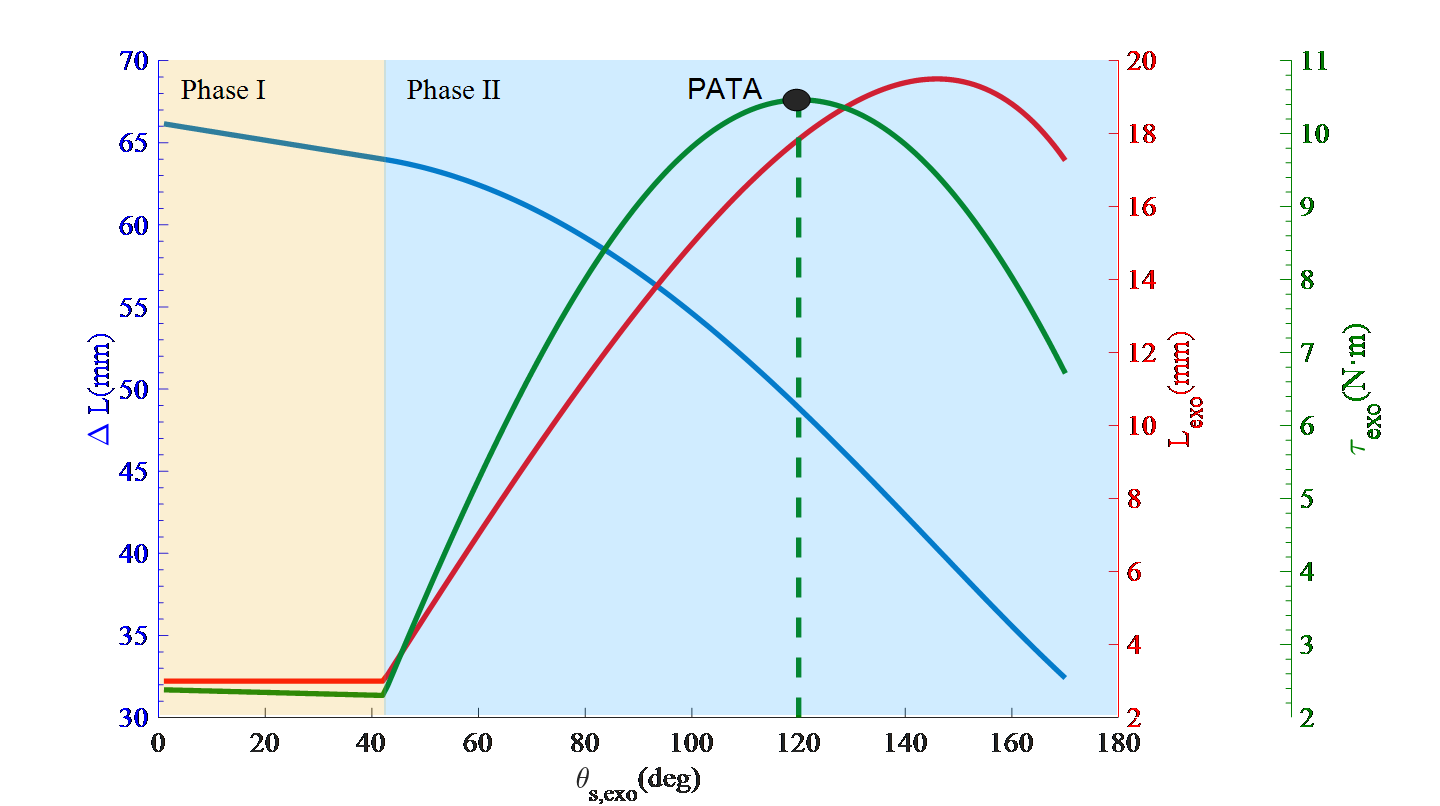}\label{fig7b}}
	\caption{ The principal of the assistive torque profile. (a) Overlapped location configurations of the force arm \(L_{exo}\), \(\theta_{c}\) is the critical angle between phases $\mathrm{\uppercase\expandafter{\romannumeral1}}$ and $\mathrm{\uppercase\expandafter{\romannumeral2}}$, \(r_{2}\) is the distance between the guide rail and the sagittal F/E joint, \(\beta\) is the angle of pulley3, as shown in Fig. \ref{fig6b} Torque profile \(\tau_{exo}\) along with the force arm \(L_{exo}\)  and total extension length of the parallel spring group \(\Delta{L}\). The area shaded in light orange is Phase $\mathrm{\uppercase\expandafter{\romannumeral1}}$, and that in light blue is Phase $\mathrm{\uppercase\expandafter{\romannumeral2}}$.}
	\label{fig7}
\end{figure*}

\begin{figure}[!t]
	\centerline{\includegraphics[width=\columnwidth]{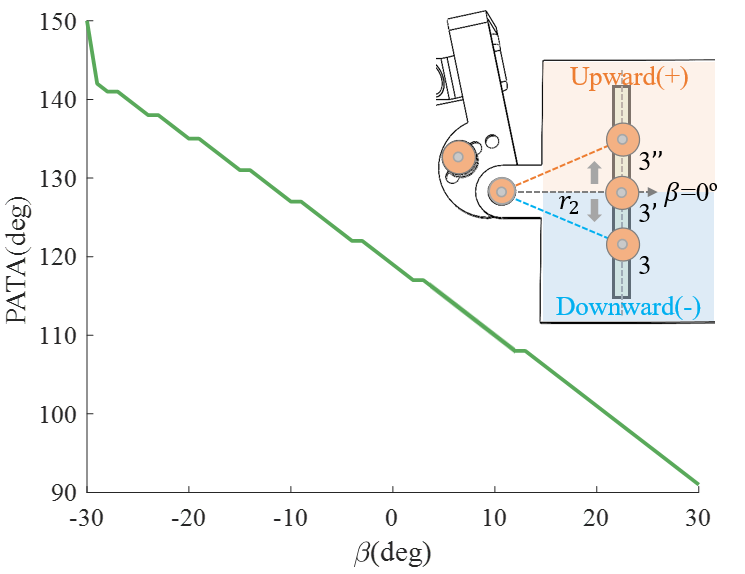}}
	\caption{The relationship between \(\beta\) and PATA.}
	\label{fig8}
\end{figure}

\subsection{Torque Generator with Adjustable PATA}
The torque generator is composed of a sagittal F/E joint bar, PATA adjus module, parallel spring group, spring pretension tuning module, carbon fiber shell, and five pulleys with inner bearings (Fig. \ref{fig6a}). A wire rope of 1mm diameter connects the parallel spring group to pulley 2. The pulleys 3 and 4 can be moved vertically through the guide rail, resulting in the adjustable PATA. The 8-shaped loop with a rotational DoF connects the parallel spring group with the spring pretension tuning module. It ensures that the parallel spring group does not experience axial rotation when screwing the spring pretension tuning module. Three identical springs (i.e., \(L_{s}\)=\(L_{AB}\)=\(L_{CD}\)=\(L_{EF}\)) are installed in the parallel spring group to obtain a high energy storage capability.

The assistive principle is to store or release the elastic potential energy of springs as the arm is lowered or raised, respectively. Since the torque generator follows the arm movement via an arm sleeve, it can be considered that the sagittal F/E angle of the exoskeleton (\(\theta_{s,exo}\)) is approximately equal to that of the user (\(\theta_{s,user}\)). The generated assistive torque ($\tau_{exo}$) can be formulated as:
\begin{equation}\tau_{exo}=F_{s}L_{exo}.\label{eq1}\end{equation}
where \(F_{s}\) is the tension generated by springs, and \(L_{exo}\) denotes the perpendicular distance between point O and the wire rope (Fig. \ref{fig7a}). Furthermore, \(F_{s}\) can be expressed as:
\begin{equation}F_{s}=K\Delta{L}.\label{eq2}\end{equation}
where \(K\) is the stiffness of the parallel spring group, which is triple that of a single spring (\(k_{s}\)), i.e., \(K\)=3\(k_{s}\), and \(\Delta{L}\) represents the total extension length of the parallel spring group. Fig. \ref{fig7b} displays the generated assistive torque (\(\tau_{exo}\)) under a set of design parameters (can be found in Table \ref{tab3}), along with \(L_{exo}\) and \(\Delta{L}\). To derive the assistive torque, the total extension length of the parallel spring group (\(\Delta{L}\)) should be formulated as a function of the sagittal F/E angle of the exoskeleton (\(\theta_{s,exo}\)). During the shoulder movement, the extension length of the parallel spring group can be considered equivalent to the length change of the wire rope between points A' and B', as they are connected in series. Therefore, the total extension length of the parallel spring group (\(\Delta{L}\)) can be expressed as:
\begin{equation}
	\Delta{L}(\theta_{s,exo}) = L_{A'B'}(\theta_{s,exo}) - L_{A'B'}(\max(\theta_{s,exo})) + L_{i}.
	\label{eq3}
\end{equation}
where \(L_{A'B'}\) is the length of the wire rope routed from A' to B' (Fig. \ref{fig6b} and \ref{fig6c}). A' and B' are the starting and ending points of the wire rope that attach to pulley 2 and pulley 4, respectively, and \(L_{i}\) denotes the initial extension length caused by the spring pretension tuning module. The minimum value of \(L_{A'B'}\) is obtained at the maximum value of \(\theta_{s,exo}\), which is set to 170º due to the safety limit.

Based on (\ref{eq1}) to (\ref{eq3}) and Fig. \ref{fig7b}, the main factor affecting PATA is \(L_{exo}\), rather than \(\Delta{L}\), as it is monotonically decreasing. The primary parameters affecting the phase shift of \(L_{exo}\) are \(\alpha\) and \(\beta\), denoting the angles of pulleys 2 and 3, as shown in Fig. \ref{fig6b} and \ref{fig6c}. As shown in Fig. 7 (a) and (b), when  \(\alpha\)  decreases or  \(\beta\)  increases, PATA would decrease and thus the phase of  \(L_{exo}\) curve would shift to the left. The assistive torque would be influenced by \(L_{exo}\) and also have a phase shift (as shown in Fig.\ref{fig1}). To better investigate the trend of PATA, when \(\alpha\)=\(\beta\)=0, we defined the PATA as a reference, denoted by \(\theta_{0}\). The PATA (\(\theta_{PATA}\)) can be derived as:
\begin{equation}\theta_{PATA}=\theta_{0}+\alpha-k\beta.\label{eq4}\end{equation}
\begin{equation}\theta_{0}=90°-arctan(\frac{r}{r_{2}}).\label{eq5}\end{equation}
where \textit{r} is the radius of pulley 3, which is 7mm in this paper, thus \(\theta_{0}\) is 80º; and \textit{k} is a variable positive coefficient around 1, due to the nonlinear effect of \(\Delta{L}\) on \(\tau_{exo}\). To change the PATA for adapting different overhead tasks, we adjusted \(\beta\) by  moving pulleys 3 and 4 along the guide rail, the relationship between \(\beta\) and PATA can be found in Fig. \ref{fig8}. We employed \(\alpha\) to control the maximum sagittal F/E joint angle (\(\theta_{s,exo}\)). \(\alpha\) can be adjusted by changing the mounting position of pulley 2 through the pre-drilled holes, and its mappings to the maximum of \(\theta_{s,exo}\) are indicated by the dotted line in Fig. \ref{fig9}e.

\subsection{Parameters Impact on Torque Profiles}

In this subsection, we analyzed the impact of design parameters on the peak value and PATA of the torque profile and conducted a simulation of torque profiles. Table \ref{tab3} shows the parameters of the torque generator.

\begin{figure}[!t]
	\centerline{\includegraphics[width=\columnwidth]{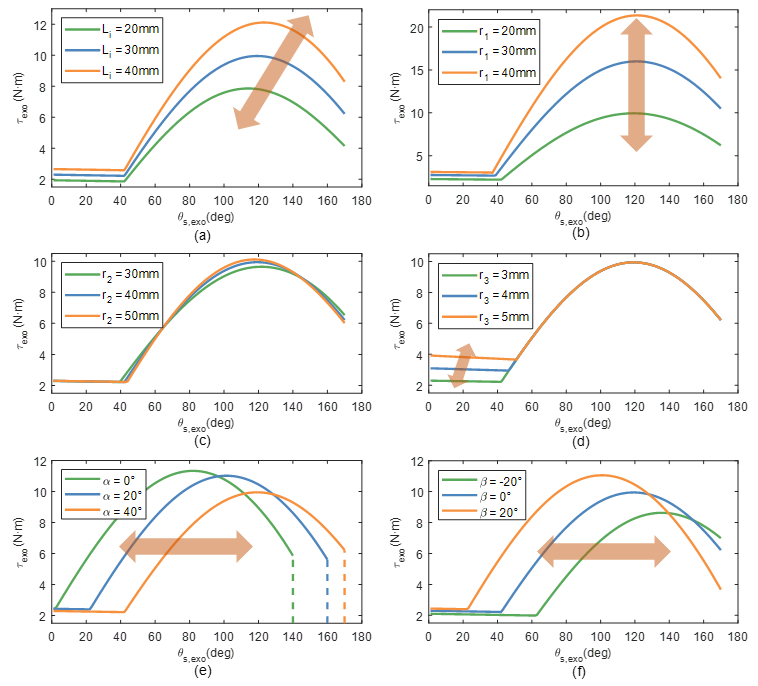}}
	\caption{The simulated assistive torque (\(\tau_{exo}\)) profiles were obtained via tuning of the assistive parameters. (a) \(L_{i}\), (b) \(r_{1}\), (c) \(r_{2}\), (d) \(r_{3}\), (e) \(\alpha\), (f) \(\beta\).}
	\label{fig9}
\end{figure}

As visualized in Fig. \ref{fig6b} and \ref{fig6c}, the generated assistive torque can be divided into two phases based on whether the wire rope is routed to the sagittal F/E joint bar. When the wire rope is attached to the sagittal F/E joint bar (Phase $\mathrm{\uppercase\expandafter{\romannumeral1}}$, Fig. \ref{fig6b}), the assistive torque remains low to avoid hindering normal activities like walking, corresponding to the low assistance zone (Fig. \ref{fig7b}). In contrast, when the wire rope is detached from the sagittal F/E joint bar (Phase $\mathrm{\uppercase\expandafter{\romannumeral2}}$, Fig. \ref{fig6c}), the assistive torque increases rapidly to provide effective assistance for the target task, corresponding to the high assistance zone (Fig. \ref{fig7b}). We defined the critical angle for switching between phases $\mathrm{\uppercase\expandafter{\romannumeral1}}$ and $\mathrm{\uppercase\expandafter{\romannumeral2}}$ as \(\theta_{c}\) (Fig. \ref{fig7a}), which is dependent on \(\alpha\) and \(\beta\) (Fig. \ref{fig6b} and \ref{fig6c}). As \(\alpha\) increases or \(\beta\) decreases, \(\theta_{c}\) increases (Fig. \ref{fig9}e - \ref{fig9}f). The wire rope coupled with the sagittal F/E joint bar formulates a passive clutch mechanism that allows for switching between low and high assistance zones.If there is no sagittal F/E joint bar (i.e., orange dotted line in Fig. \ref{fig6b}), the torque generator will generate torque to pull the arm backward, which poses a safety hazard to the user.

\begin{table}
	\centering
	\caption{THE EFFECTS OF PARAMETERS (\(\alpha\), \(\beta\), \(r_{1}\), \(r_{2}\), \(r_{3}\), \(L_{i}\)) ON TORQUE ASSISTIVE AND PATA IN BOTH PHASES}
	\setlength{\tabcolsep}{3pt}
	\begin{tabular}{>{\centering\arraybackslash}p{60pt}>{\centering\arraybackslash}p{50pt}>{\centering\arraybackslash}p{50pt}}
		\toprule
		Requirement& Parameter& Value \\
		\midrule
		\multirow{4}{*}{Peak torque} & K &12N/mm \\
		& \(L_{i}\) & 30mm \\
		& \(r_{1}\) & 20mm\\
		& \(r_{2}\) & 40mm \\
		
		Torque at phase $\mathrm{\uppercase\expandafter{\romannumeral1}}$ & \(r_{3}\) & 3mm \\
		Critical angle & \(\alpha\)& 40°\\
		\bottomrule
	\end{tabular}
	\label{tab3}
\end{table}

\begin{table}
	\centering
	\caption{THE EFFECTS OF PARAMETERS (\(\alpha\), \(\beta\), \(r_{1}\), \(r_{2}\), \(r_{3}\), \(L_{i}\)) ON TORQUE ASSISTIVE AND PATA IN BOTH PHASES}
	\setlength{\tabcolsep}{3pt}
	\begin{tabular}{>{\centering\arraybackslash}p{50pt}>{\centering\arraybackslash}p{50pt}>{\centering\arraybackslash}p{50pt}>{\centering\arraybackslash}p{50pt}}
		\toprule
		\multirow{2}{*}{Parameter} & Phase $\mathrm{\uppercase\expandafter{\romannumeral1}}$ & \multicolumn{2}{c}{Phase $\mathrm{\uppercase\expandafter{\romannumeral2}}$} \\
		& \(\tau_{exo}\) & \(\tau_{exo}\) & PATA \\
		\midrule
		\(\alpha\) & \(\mathrm{NI^{a}}\) & \(\mathrm{N^{b}}\) & \(\mathrm{P^{c}}\) \\
		\(\beta\) & NI & P & N \\
		\(r_{1}\) & NI & P & NI\\
		\(r_{2}\) & NI& NI& NI\\
		\(r_{3}\) & P& NI& NI\\
		\(L_{i}\) & P& P& NI\\
		\bottomrule
		\multicolumn{3}{p{180pt}}{$^{\mathrm{a}}$: No Impact; $^{\mathrm{b}}$: Negative; $^{\mathrm{c}}$: Positive.}
	\end{tabular}
	\label{tab4}
\end{table}

\(L_{exo}\) and \(L_{A'B'}\) can be geometrically calculated in the two phases. Based on (\ref{eq1}) to (\ref{eq3}) , the peak value of the assistive torque increases with the stiffness (K) and initial extension length (\(L_{i}\))  of the parallel spring group in both phases. In phase $\mathrm{\uppercase\expandafter{\romannumeral1}}$, the primary parameter affecting \(L_{exo}\) and \(L_{A'B'}\) is \(r_{3}\), the radius of the sagittal F/E joint bar. As \(r_{3}\) increases, \(L_{exo}\) and \(L_{A'B'}\) increase, resulting in an increase in assistive torque within the low assistance zone. In phase $\mathrm{\uppercase\expandafter{\romannumeral2}}$, the primary parameters affecting \(L_{exo}\) and \(L_{A'B'}\) are \(\alpha\), \(\beta\), \(r_{1}\), where \(r_{1}\) represents the distance between pulley 2 and the sagittal F/E joint. As \(r_{1}\) increases, \(L_{exo}\) and \(L_{A'B'}\) increase, causing an increase in the peak value of the assistive torque. When \(\alpha\) decreases or \(\beta\) increases, \(L_{A'B'}\) increases, leading to an increase in the peak value of the assistive torque, and the impact on \(L_{exo}\) has been stated in subsection C. We also extracted and tested \(r_{2}\), the distance between the guide rail and the sagittal F/E joint, as a design parameter. The result showed that it essentially has no significant impact on the peak or phase of the assistive torque, allowing us to choose an appropriate value according to the requirements of the design space. Table \ref{tab4} summarizes the effects of the parameters (\(\alpha\), \(\beta\), \(r_{1}\), \(r_{2}\), \(r_{3}\), \(L_{i}\)) on the assistive torque and PATA.

Fig. \ref{fig9} demonstrates how the parameters affect the assistive torque profiles through simulations with an ideal situation (i.e., no deformation and friction). The simulation results in Fig. \ref{fig9} are in accord with our previous analyses. In particular, the parameters \(L_{i}\), \(\alpha\) and \(\beta\) can be easily adjusted, and  K can be easily changed by replacing the spring. Consequently, this implies that the main characteristics of the torque profile can be tuned by reconfiguring the exoskeleton (i.e. by tuning the pre-tension of springs or the mounting position of pulley 2), rather than reconstructing it. As far as we know, this is the first exoskeleton that can simultaneously modify these characteristics of the torque profile without the need of reconstruction.

\subsection{PUEI and Size Regulation Module}

The PUEI contains a lower back linkage and all components that attach the exoskeleton to the user, e.g., shoulder straps, chest buckles, and waist band. These flexible fabrics ensure a compliant human-exoskeleton interface. Furthermore, the lower back linkage is made of aluminum alloy, which transmits the force compensated by the torque generator to the waist.

The size regulation module aims to meet different ergonomic requirements and guarantee comfort and safety. The height and width of the back can be controlled with the upper and lower back linkages and lateral back slides, respectively. The flexible fabric can be adjusted to tighten the exoskeleton to the waist, back, and arm of the user.

\section{EXPERIMENTAL EVALUATIONS}
\label{sec3}
Two experimental sessions were conducted to verify the HIT-POSE has (a) a sufficient ROM (session 1), and (b) an ability and adaptability to assist different overhead tasks (session 2). 

\subsection{Participants}

Two separate groups of participants were recruited for this study. Five healthy subjects participated in session 1 (all male, all right-handed, age: 24.1 ± 1.64 years, height:  173.8 ± 3.27 cm, weight: 70 ± 9.06 kg), and ten healthy subjects participated in session 2 (all male, all right-handed, age: 25.1 ± 2.0 years, height: 177 ± 4.3 cm, weight: 68.3 ± 9.3 kg). All participants had no prior experience with the exoskeleton and signed an informed consent before the experiment. The experimental protocol received approval from the Chinese Ethics Committee of Registering Clinical Trials (ChiECRCT20200319).

\subsection{Data Recordings}

ROM of the right-side shoulder was measured by the motion capture system (100Hz, VICON, Oxford Metrics, Oxford, UK) to verify whether the exoskeleton can provide sufficient ROM. Muscular activities were monitored using the surface EMG electrodes (1111Hz, Delsys Inc., Natick, MA, USA) to explore the impact of the exoskeleton on biomechanical effort reduction. Meanwhile, acceleration signals were recorded through an IMU (148Hz, Delsys Inc., Natick, MA, USA). Notably, the motion capture system did not work simultaneously with EMG. Thus, we did not synchronize them through a trigger.

\begin{figure}[!t]
	\centerline{\includegraphics[width=\columnwidth]{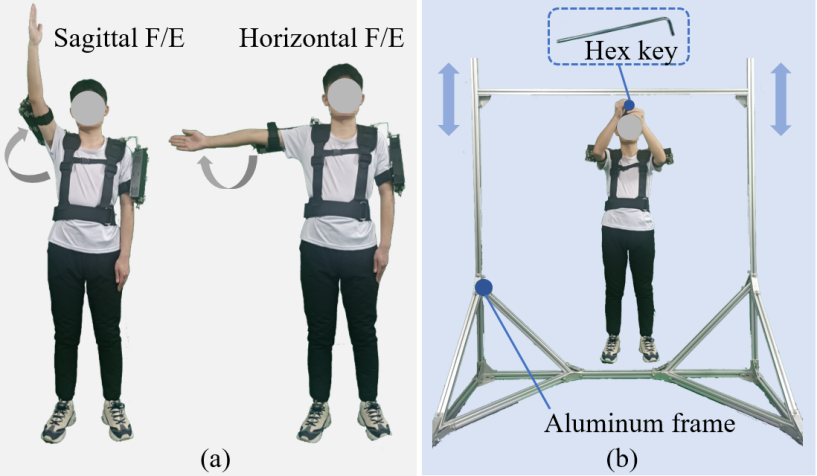}}
	\caption{Experimental tasks. (a) Shoulder movements in session 1. (b) Screwing task in session 2.}
	\label{fig10}
\end{figure}

\begin{figure}[!t]
	\centerline{\includegraphics[width=\columnwidth]{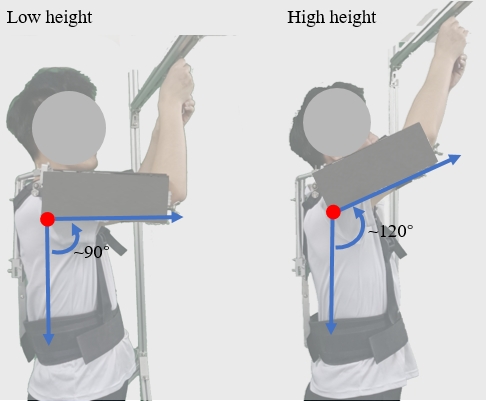}}
	\caption{The definition of the low (left) and high (right) heights.}
	\label{fig11}
\end{figure}

\subsection{Experimental Tasks}
To verify that HIT-POSE can provide sufficient ROM (session 1), participants were directed to maximize sagittal and horizontal F/E movements, at a self-selected constant pace, as shown in Fig. 10(a). For validating the performance of HIT-POSE (session 2), participants were guided to screw the M6 bolts with a hex key (mass: 8.6 g), as shown in Fig. 10(b). Nuts were embedded in an aluminum frame, which was set to two heights: low height was defined as hand with the shoulder at about 90º, and high height was defined as hand with the shoulder at about 120º (Fig. \ref{fig11}).

\subsection{Procedure}

Prior to the start of the experiment, participants were introduced to the experimental paradigm and guided to familiarize HIT-POSE and tasks. In session 1, tasks were performed under two conditions: without the exoskeleton (NO HIT-POSE) and with the exoskeleton (HIT-POSE). Each participant was required to complete 6 trials (3 trials were repeated in each condition). In session 2, the exoskeleton had two configurations for screwing at low and high heights, respectively (Fig. 12). The two configurations present the same peak torque but different PATAs corresponded to the heights. The condition where the PATA of the exoskeleton aligned with the PATA required for the height, was referred to as HIT-POSE Match. Otherwise, it was defined as HIT-POSE Mismatch. Each task was performed in 3 conditions: NO HIT-POSE, HIT-POSE Match, and HIT-POSE Mismatch (Fig. 12).

Surface EMG electrodes were placed unilaterally on the right-side muscles according to SENIAM guidelines \cite{hermens2000development}: anterior deltoid (AD), middle deltoid (MD), posterior deltoid (PD), bicep brachii (BB), pectoralis major (PM), trapezius (TR), latissimus dorsi (LD), erector spinae (ES). Additionally, an IMU was placed on the right wrist to measure the acceleration. Then, the maximum voluntary contraction (MVC) of each muscle was measured. The order of the conditions and tasks was randomized across participants to avoid order effects. The NASA Task Load Index (NASA-TLX) \cite{hart2006nasa} and Borg CR-10 questionnaires \cite{Borg} were collected after each trial, to assess the perceived workload (PW) and RPE, respectively. After all trials wearing the exoskeleton ended, the participants filled out the System Usability Scale (SUS) \cite{brooke1996sus}. Fig. 12 presents the experimental procedure for session 2.

\begin{figure}[!t]
	\centerline{\includegraphics[width=\columnwidth]{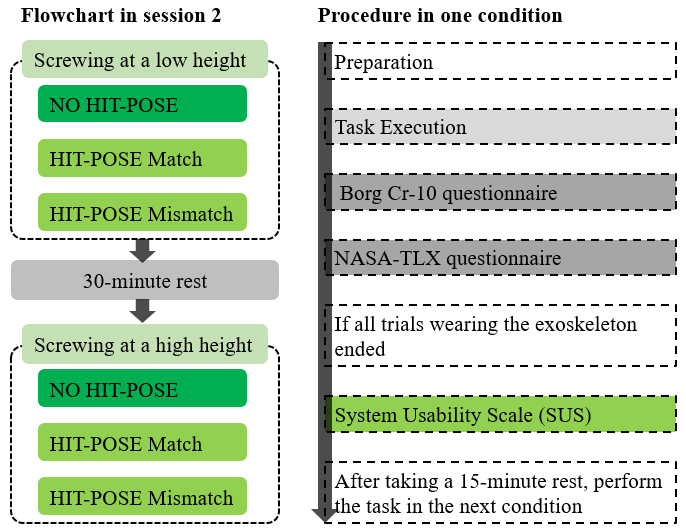}}
	\caption{The experimental flowchart (left) in session 2 and procedure in one condition (right). The order of the conditions and tasks is randomized.}
	\label{fig12}
\end{figure}

\subsection{Data and Statistical Analysis}

In session 1, shoulder horizontal and sagittal F/E angles were obtained through a standardized data processing pipeline in VICON. Mean values of the ROM were calculated among all participants in NO HIT-POSE and HIT-POSE conditions. In session 2, acceleration signals were resampled to 1111Hz and variances were calculated through a 30ms time sliding window. The start and end times of the task for each trial were determined by a given variance threshold. EMG signals were processed to compute the linear envelopes by setting them to zero-mean, applying band-pass filter (4th, 20-350Hz), full-wave rectification, low-pass filter (6Hz), and normalization. For each trial, the root mean square (RMS) of each muscle was obtained from the linear envelope via a 30ms time sliding window during the phase of task execution.

Statistical analyses were used to assess the effectiveness of the exoskeleton in reducing muscle activations. The Shapiro-Wilk test was employed to check data normality. If the normal distribution was not satisfied, the Wilcoxon signed-rank test was used to perform statistical analysis, otherwise, the t-test was utilized. Statistical significance was concluded when \textit{p} \(<\) 0.05.

\section{RESULTS}
\label{sec4}
\subsection{ROM}
Table \ref{tab5} reports measured ROM averaged among all participants in NO HIT-POSE and HIT-POSE conditions. No significant differences were observed in maximum sagittal and horizontal F/E angle between conditions ( \textit{p} = 0.6432 and \textit{p} = 0.7421, respectively).

\begin{table}
	\centering
	\caption{MEASURED ROM WITHOUT AND WITH HIT-POSE}
	\setlength{\tabcolsep}{3pt}
	\begin{tabular}{>{\centering\arraybackslash}p{50pt}>{\centering\arraybackslash}p{70pt}>{\centering\arraybackslash}p{70pt}}
		\toprule
		Condition& Maximum sagittal F/E angle (deg) &Maximum horizontal F/E angle (deg)\\
		\midrule
		NO HIT-POSE& 165.13 ± 2.15& 159.74 ± 6.27 \\
		HIT-POSE& 164.46 ± 2.21& 158.28 ± 7.20 \\
		\bottomrule
	\end{tabular}
	\label{tab5}
\end{table}

\subsection{Muscular Activations}
We evaluated the impact of the exoskeleton on muscular activations, by comparing wearing and not wearing the exoskeleton and comparing Match and Mismatch conditions. Herein, “wearing the HIT-POSE” refers to both HIT-POSE Match and HIT-POSE Mismatch conditions. As shown in Fig. 13, when wearing the HIT-POSE to perform both tasks, the AD, MD, PD, BB, TR, and ES muscles exhibited significantly decreased EMG activations compared to not wearing the HIT-POSE. In HIT-POSE Match conditions for both tasks, significantly decreased muscular activations in all monitored muscles were found compared to NO HIT-POSE and HIT-POSE Mismatch conditions. In HIT-POSE Mismatch conditions for both tasks, compared to the NO HIT-POSE condition, there is a significant reduction in muscle activation for all monitored muscles other than PM and LD.

 As shown in Table \ref{tab6}, in HIT-POSE Match conditions for both tasks, the absolute percentage reduction was above 15\% (highest at 22.9\% and 23.8\% for AD in screwing at low and high heights, respectively), and the relative percentage reduction was above 30\% (highest at 48.3\% for AD in screwing at a low height and 49.6\% for BB in screwing at a high height). In HIT-POSE Mismatch conditions for both tasks, the absolute percentage reduction was above 5\% (highest at 11.9\% for AD in screwing at a low height, and 11.3 for BB and TR in screwing at a high height), and the relative percentage reduction was above 9\% (highest at 24.3\% for AD in screwing at a low height and 27.7\% for BB in screwing at a high height). For the same muscle in the same task, the absolute and relative decrease percentages in the HIT-POSE Match condition are approximately twice as much as in the HIT-POSE Mismatch condition.

\begin{table}
	\centering
	\caption{REDUCTION PERCENTAGE ($\%$) OF MUSCLE ACTIVATIONS WHEN WEARING HIT-POSE.}
	\setlength{\tabcolsep}{3pt}
	\begin{tabular}{>{\centering\arraybackslash}p{20pt}>{\centering\arraybackslash}p{50pt}>{\centering\arraybackslash}p{50pt}>{\centering\arraybackslash}p{50pt}>{\centering\arraybackslash}p{50pt}}
		\toprule
		\multirow{2}{*}{\vspace*{-3mm}Muscles\vspace*{0mm}} & \multicolumn{2}{c}{Screwing at a low height} & \multicolumn{2}{c}{Screwing at a high height} \\
		& HIT-POSE Match ($\%$) & HIT-POSE Mismatch ($\%$) & HIT-POSE Match ($\%$) & HIT-POSE Mismatch ($\%$) \\ 
		\midrule
		AD&	22.9$|$48.3&	11.5$|$24.3&	23.8$|$48.5&	10.6$|$21.6 \\
		MD&	15.6$|$30.3&	6.5$|$12.5&	19.5$|$37.9&	8.2$|$15.8 \\
		PD&	18.5$|$35.6&	7.0$|$13.5&	23.3$|$44.7&	8.9$|$17.1 \\
		BB&	15.5$|$41.9&	7.6$|$20.6&	20.1$|$49.6&	11.3$|$27.7 \\
		PM&	19.0$|$43.4&	5.5$|$12.5&	17.9$|$31.9&	5.4$|$9.6 \\
		TR&	19.6$|$40.9&	8.8$|$18.2&	20.2$|$42.2&	11.3$|$23.6 \\
		LD&	17.4$|$47.0&	7.6$|$20.6&	15.7$|$42.5&	7.2$|$19.6 \\
		ES&	15.9$|$43.3&	8.0$|$21.8&	17.0$|$43.6&	8.3$|$21.4 \\
			
		\bottomrule
		\multicolumn{5}{p{250pt}}{A$|$B: A and B denote absolute and relative percentage reduction with respect to NO HIT-POSE condition, respectively.}
	\end{tabular}
	\label{tab6}
\end{table}

\begin{figure}[!t]
	{\includegraphics[width=\columnwidth]{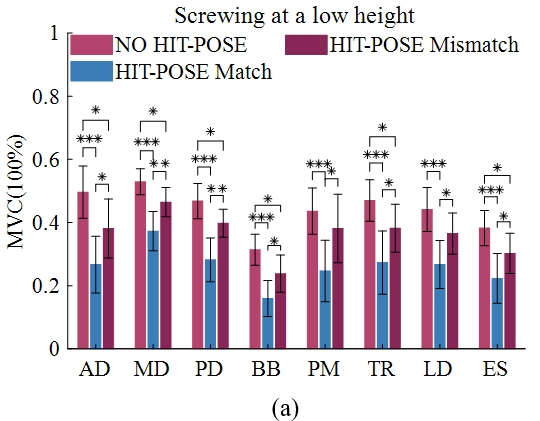}\label{fig13a}}
	{\includegraphics[width=\columnwidth]{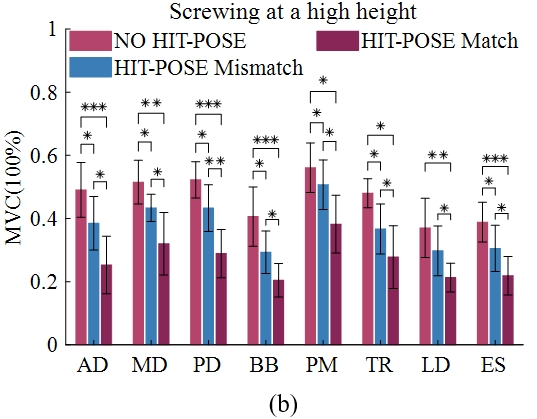}\label{fig13b}}
	\caption{Muscular activations within different conditions for two tasks from pairwise comparisons. (a) Screwing at a low height. (b) Screwing at a high height. The error bar represents the mean ± standard deviation. *:  \textit{p} \(<\) 0.05; **: \textit{p} \(<\) 0.01; ***: \textit{p} \(<\) 0.001.}
	\label{fig13}
\end{figure}

\subsection{Subjective Feedback}

Fig. 14 shows the RPE scores averaged across participants. In HIT-POSE Match conditions for both tasks, RPE scores showed a significant difference relative to the other two conditions. In HIT-POSE Mismatch conditions for both tasks, no significant differences were observed compared to the NO HIT-POSE condition. NASA-TLX questionnaire assesses PW across six aspects. As indicated in Fig. 15, in HIT-POSE Match conditions for both tasks, significant differences were found in physical demand, effort, and frustration, compared to the other two conditions. No significant differences were observed in mental demand, temporal demand, and performance. In addition, the average SUS score was 79.7 ± 5.9.

\begin{figure}[!t]
	\centerline{\includegraphics[width=\columnwidth]{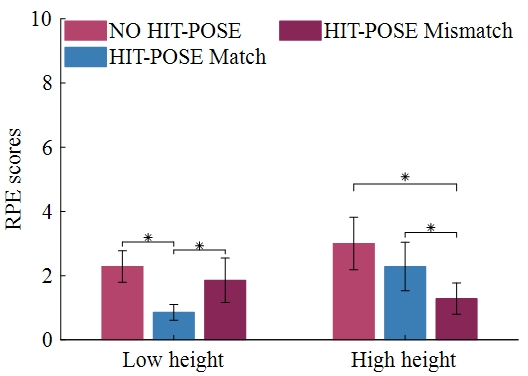}}
	\caption{RPE scores for each task among the three conditions from pairwise comparisons. The error bar represents the mean ± standard deviation. *:  \textit{p} \(<\) 0.05; **: \textit{p} \(<\) 0.01; ***: \textit{p} \(<\) 0.001.}
	\label{fig14}
\end{figure}

\begin{figure}[!t]
	{\includegraphics[width=\columnwidth]{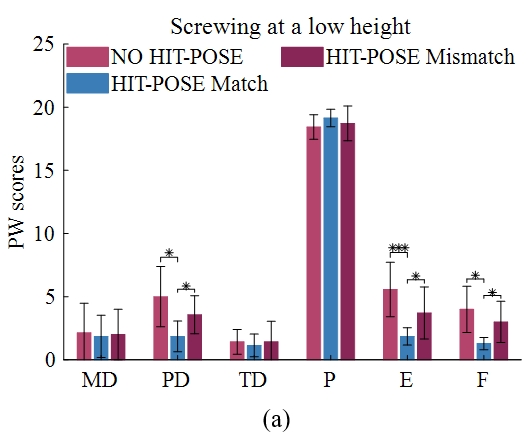}\label{fig15a}}
	{\includegraphics[width=\columnwidth]{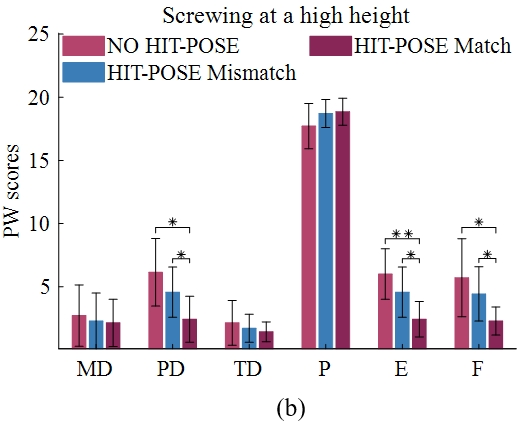}\label{fig15b}}
	\caption{PW scores for each task among the three conditions from pairwise comparisons. (a) Screwing at a low height. (b) Screwing at a high height. MD = mental demand, PD = physical demand, TD = temporal demand, P = performance, E = effort, F = frustration. The error bar represents the mean ± standard deviation. *:  \textit{p} \(<\) 0.05; **: \textit{p} \(<\) 0.01; ***: \textit{p} \(<\) 0.001.}
	\label{fig15}
\end{figure}

\section{DISCUSSION}
\label{sec5}
In this study, we propose a novel passive occupational shoulder exoskeleton with adjustable PATA for different overhead tasks. We first designed an ergonomic shoulder structure through kinematic modeling and ROM simulations to ensure a sufficient shoulder ROM. Then we implemented a torque generator with pulleys moved along the guide rail, thus realizing adjustable PATA. Finally, we also investigated parameters' impact on torque profiles, providing users with valuable recommendations for achieving various torque profiles through the considerable reconfiguration flexibility of the exoskeleton to adapt to diverse tasks and individuals.

\subsection{Kinematics Compatibility}
The HIT-POSE is equipped with an ergonomic shoulder structure to ensure sufficient ROM for the user. This implies favorable kinematics compatibility, which has a profound impact on the exoskeleton’s usability and user acceptance. Typically, some passive shoulder exoskeletons integrate a multi-DoF chain or simplified two-link hinge to increase the ROM, such as MATE \cite{pacifico2023evaluation} and PAEXO \cite{schmalz2019biomechanical}. However, they may collide with nearby objects easily due to the large volume and protrusions. Recently, a similar study has proposed a shoulder structure to assist overhead tasks in the shipbuilding industry \cite{kim2022passive}. However, the ROM of the horizontal F/E is limited, and the design parameters are fixed, thus lacking reconfigurability. This reduces the adaptability and increases the manufacturing costs of the exoskeleton.

To ensure sufficient ROM while compactness, we extracted three parameters (\(\phi\), \(d_{v}\), \(d_{b}\)) to design the shoulder structure, and two of them (\(d_{v}\) and \(d_{b}\)) an be adjusted without reconstructing the exoskeleton. A suitable set of parameters (\(\phi\)=15°, \(d_{v}\)=80mm, \(d_{b}\)=10mm) was selected through kinematic modeling and ROM simulations.Sufficient ROM was guaranteed in both sagittal and horizontal F/E for performing overhead tasks and natural movements, as shown in Table \ref{5}, while the structure with a relatively small volume is designed to reduce the possibility of collision.

\subsection{Comparison of Adaptability with Existing Exoskeletons}
Different overhead tasks usually have different PATAs and thus different optimal torque profiles \cite{maurice2019objective, hyun2019light}. Additionally, when handling the same task, individuals with different heights may have different target angles, i.e., different tasks. Consequently, the task and individual adaptability of the exoskeleton are crucial for users, which essentially involves adjustable assistive profiles, including peak torque and PATA. Existing exoskeletons have made considerable efforts in adjusting the assistive levels with different peak torque.  Several exoskeletons employ a manual knob or an electrical component to adjust the assistive level, such as the Shoulder X \cite{pinho2022shoulder} and H-PULSE \cite{grazi2022kinematics, grazi2020design}, other devices require to replace the spring or other parts affecting the spring length. However, existing exoskeletons can only adjust the amplitude of peak torque, thus just adapting to a specific task. Employing torque profiles with a fixed PATA to cope with all different tasks and individuals is not an energy-efficient way and limits the adaptability of the torque generator \cite{hyun2019light}, \cite{vazzoler2022analysis}. For such exoskeletons, if we need to adjust the PATA, we can only change it by reconstructing the exoskeleton.

To enhance the task adaptability and personalized customization capability of the exoskeleton, we proposed a torque generator with an adjustable PATA. We utilized a set of parameters (\(\alpha\), \(\beta\), \(r_{1}\), \(r_{2}\), \(r_{3}\), \(L_{i}\)) to guide the design of the torque generator, and investigated their impacts on torque profiles via geometric analysis and simulations. A subset of parameters (\(\alpha\), \(\beta\), \(L_{i}\)) can be adjusted to reconfigure the exoskeleton, each controlling peak torque, PATA, and safety limit angle, respectively. Notably, \(\beta\) can be adjusted within a range of -30º to 30º, resulting in a range of PATA from 90º to 150º, which meets the requirements of various overhead tasks and individuals. The experimental results demonstrated that monitored muscular activations in HIT-POSE Match condition decreased the most compared to the NO HIT-POSE condition, thus validating the effectiveness of PATA.

\subsection{Effects of the Exoskeleton on Muscular Activations}

To provide a comprehensive overview of the effects of the HIT-POSE on muscular activations, we monitored and analyzed eight muscles primarily involved in shoulder movement and stability in three conditions (i.e., NO HIT-POSE, HIT-POSE Match, and HIT-POSE Mismatch) during screwing at low and high heights. Overall, our findings indicated that the HIT-POSE can significantly reduce muscular activations of related muscles, thus potentially mitigating the incidence of shoulder WMSDs. 

When wearing the exoskeleton in both tasks, significantly reduced muscular activations were observed in agonist muscles (AD, MD, BB, TR) and antagonist muscles (PD). The exoskeleton provided support to the human shoulder, allowing agonist muscles to generate less force to achieve the same action, while also decreasing co-contraction of the antagonist muscles \cite{grazi2020design}. Similar results can be found in the ShoulderX \cite{van2018evaluation}, PAEXO \cite{maurice2019evaluation}, MATE \cite{pacifico2020experimental}, and H-VEX \cite{hyun2019light}. This observation reflects the rational and effective design of the device. As reported by a previous study \cite{lee2002dynamic}, the reduced muscular activation implies lower force, resulting in decreased compressive force on the shoulder, and thus potentially reducing biomechanical stress on the shoulder girdle. Such positive effects suggest that long-term use of the exoskeleton has the potential to reduce the incidence of WMSDs, as reported by previous studies \cite{de2023passive, de2022occupational}. Additionally, for both tasks, PM (agonist muscle) and LD (antagonist muscle) demonstrated significantly reduced muscular activations in the HIT-POSE Match condition, but not in the HIT-POSE Mismatch condition. The difference is likely due to the fact that less assistance was provided in the HIT-POSE Mismatch condition, or individuals adopted different shoulder postures for different tasks as well as conditions. Nevertheless, the more accurate explanations remain to be explored further. 

Notably, a reduction in the ES muscle was also expected when wearing the HIT-POSE in both tasks. The result is consistent with the existing study \cite{jakobsen2023biomechanical}. This may be attributed to the fact that the exoskeleton transmits forces to the waist through the device's back bar and waist belt. The result could indicate that the HIT-POSE doesn’t increase the risk of back WMSDs. As expected, for both tasks, all muscular activations decreased the most in the HIT-POSE Match condition. This phenomenon can be explained by the fact that matching the PATA with the target task angle results in increased assistive torque, with reductions consistent with the level of assistance \cite{grazi2020design}. This also demonstrates the necessity of PATA for effective assistance.

\subsection{Effects of The Exoskeleton on Subjective Feedback}
In this study, we also investigated the user experience with the exoskeleton, by quantifying the changes in RPE, RPD, and PW among conditions. The physical demand and effort items in the NASA-TLX and RPE among three conditions presented similar phenomenon as the EMG results. In the HIT-POSE Match condition, they displayed statistical significance from the other two conditions, and the participants perceived the greatest assistance and thus exerted the lowest physical effort. The outcomes are in line with the previous study \cite{grazi2020design}.

In addition, scores of the frustration item in NASA-TLX also exhibited a significant reduction with HIT-POSE conditions, which is in line with the existing study \cite{grazi2020design}. Evidence may be available in a SUS question (9. I felt very confident using the system.) with a high average score (4.56 ± 0.53). We speculate that mental confidence bridges the frustration from the task, which may be related to a placebo effect. Meanwhile, the SUS score (79.7 ± 5.9) expressed good usability of the exoskeleton according to the empirical evaluation \cite{bangor2008empirical}.

\subsection{Limitations of The Study}
While the experimental evaluations have shown positive effects, this study still has some limitations. Firstly, the PATA of the torque generator needs to be regulated manually. Consequently, one of our future works is to add a lightweight electrical element to adjust PATA automatically. Secondly, this experiment was conducted in a laboratory setting. Accordingly, the participants had no work experience. A field study and even the female participants would be considered in our further research. Thirdly, the long-term effect of the exoskeleton on users is not studied, which may exhibit different results. As the proposed prototype matures, we will investigate this issue, aiming to get a more complete and accurate assessment.

\section{CONCLUSION}
In this study, we propose a passive occupational shoulder exoskeleton named HIT-POSE for overhead tasks, which enhances the adaptability of tasks and individuals while ensuring sufficient shoulder ROM. Firstly, we design an ergonomic shoulder structure by kinematic modeling and ROM simulations and verify through experiments. Secondly, we implement a torque generator with adjustable PATA through geometric analysis and simulations of torque profiles. Thirdly, we investigate the parameters' impacts on torque profiles to determine adjustable parameters to achieve various torque profiles with different PATA without reconstructing the exoskeleton. Finally, we conduct experimental evaluations with aspects of objective and subjective within three conditions for two tasks, to validate the effectiveness of our proposed exoskeleton. Overall, the results have shown that HIT-POSE can decrease the physical load on users and has the potential to reduce the incidence of shoulder WMSDs.

\section*{Acknowledgment}

We gratefully acknowledge the participation of all individuals involved in this experiment. Their dedication and support were crucial to the completion of our research.

\bibliographystyle{IEEEtran}
\bibliography{IEEEabrv, myref}


\end{document}